# Preserving Privacy, Increasing Accessibility, and Reducing Cost: An On-Device Artificial Intelligence Model for Medical Transcription and Note Generation

Johnson Thomas[1], Ayush Mudgal[2], Wendao Liu[2], Nisten Tahiraj[3], Zeeshaan Mohammed[4], Dhruv Diddi[4]

## Abstract

**Background:** Clinical documentation represents a significant burden for healthcare providers, with physicians spending up to 2 hours daily on administrative tasks. Recent advances in large language models (LLMs) offer promising solutions, but privacy concerns and computational requirements limit their adoption in healthcare settings.

**Objective:** To develop and evaluate a privacy-preserving, on-device medical transcription system using a fine-tuned Llama 3.2 1B model capable of generating structured medical notes from medical transcriptions while maintaining complete data sovereignty entirely in the browser.

**Methods:** We fine-tuned a Llama 3.2 1B model using Parameter-Efficient Fine-Tuning (PEFT) with LoRA on 1,500 synthetic medical transcription-to-structured note pairs. The model was evaluated against the base Llama 3.2 1B on two datasets: 100 endocrinology transcripts and 140 modified ACI benchmark cases. Evaluation employed both statistical metrics (ROUGE, BERTScore, BLEURT) and LLM-as-judge assessments across multiple clinical quality dimensions.

**Results:** The fine-tuned OnDevice model demonstrated substantial improvements over the base model. On the ACI benchmark, ROUGE-1 scores increased from 0.346 to 0.496, while BERTScore F1 improved from 0.832 to 0.866. Clinical quality assessments showed marked reduction in major hallucinations (from 85 to 35 cases) and enhanced factual correctness (2.81 to 3.54 on 5-point scale). Similar improvements were observed on the internal evaluation dataset, with composite scores increasing from 3.13 to 4.43 (+41.5%).

**Conclusions:** Fine-tuning compact LLMs for medical transcription yields clinically meaningful improvements while enabling complete on-device browser deployment. This approach addresses key barriers to AI adoption in healthcare: privacy preservation, cost reduction, and accessibility for resource-constrained environments.

[1]Department of Endocrinology, Mercy Hospital, Springfield, Missouri, USA, [2]Starfishdata.ai, [3]alignmentlab.ai , [4]Solo Tech

# 1. Introduction

The administrative burden of clinical documentation has reached crisis proportions in modern healthcare, with physicians dedicating 50% of their time to electronic health record (EHR) tasks[1]. This documentation burden contributes significantly to physician burnout, with studies indicating that for every hour of direct patient care, clinicians spend nearly two hours on EHR-related activities [2,3]. The recent emergence of large language models (LLMs) presents opportunities to automate clinical documentation, particularly through the generation of structured clinical notes from physician-patient conversations [4].

Current AI-powered transcription solutions, while demonstrating impressive capabilities, face significant adoption barriers in healthcare settings. Cloud-based systems raise substantial privacy concerns given the sensitive nature of patient data and strict regulatory requirements under HIPAA and similar frameworks [6]. Cloud-based solutions require transmitting patient data to external servers, raising concerns about data breaches and unauthorized access. Additionally, the computational requirements and associated costs of state-of-the-art models create accessibility barriers for smaller healthcare practices and resource-constrained environments .

The advent of smaller, more efficient LLMs has opened new possibilities for on-device deployment. Models like Llama 3.2 1B demonstrate that significant language understanding capabilities can be achieved with substantially reduced computational requirements [5]. When combined with parameter-efficient fine-tuning (PEFT) techniques such as Low-Rank Adaptation (LoRA), these models can be adapted for specialized medical tasks while maintaining feasibility for local deployment [7].

This study presents the development and evaluation of an on-device medical transcription system built upon a fine-tuned Llama 3.2 1B model. Our approach addresses the critical need for privacy-preserving, cost-effective clinical documentation tools that can operate entirely on local hardware in a browser without compromising patient data security or requiring ongoing subscription costs.

## 2. Related Work

### 2.1 Medical Transcription with Large Language Models

The application of LLMs to healthcare has gained significant traction in recent years. Google's Med-PaLM and Med-PaLM 2 demonstrated that LLMs could achieve physician-level performance on medical question-answering tasks [8]. Similarly, Microsoft's BioGPT showed promising results in biomedical text generation and mining [9]. However, these models typically require substantial computational resources, limiting their deployment in resource-constrained clinical environments.

Recent work has focused on developing more compact medical LLMs. Chen et al. introduced MedAlpaca, a 7B parameter model fine-tuned on medical data [10]. While more efficient than larger models, it still requires dedicated GPU resources for real-time inference. The ChatDoctor model by Li et al. demonstrated that fine-tuning on medical conversations could improve domain-specific performance [11].The Sound of Healthcare study showed that LLMs can achieve state-of-the-art performance in medical transcription tasks, particularly in speaker diarization and Word Error Rate reduction using Chain-of-Thought prompting [12]. Other open‑source efforts have explored fine‑tuning compact LLMs (7B mistral model) for specialty‑specific note generation, demonstrating gains compared to similarly sized baselines [13]. These findings establish the technical feasibility of LLM-based medical transcription while highlighting the importance of domain-specific adaptation.

### 2.2 Privacy-Preserving Healthcare AI

Privacy preservation in healthcare AI has become a critical research focus. A commentary published in npj Digital Medicine outlined strategies including deidentification, differential privacy, synthetic data generation, and locally deployed models for better privacy [4]. The development of federated learning approaches has also enabled multi-institutional collaboration without data sharing[14].

### 2.3 On-Device AI in Healthcare

Edge computing in healthcare has gained significant traction, with systems like KidneyTalk-open no-code medical LLM deployment on desktop computers [15]. But this is only usable on a high end Mac.Recent advances in model compression and optimization have made on-device LLM deployment more feasible, with studies showing that compact models like Phi-3 Mini can achieve strong accuracy-speed balance on mobile devices, while medically fine-tuned models such as Med42 and Aloe deliver higher clinical accuracy. Notably, even older mobile devices can run these models effectively, with memory rather than processing power being the primary constraint [16].Recent developments in browser-based AI, particularly WebGPU technology, have enabled running complex models directly in web browsers. Whisper, OpenAI's speech recognition model, has been successfully ported to run in browsers using WebGPU, demonstrating the feasibility of client-side speech processing.

## 3. Methods

### 3.1 Model Architecture and Fine-tuning

We employed Llama 3.2 1B Instruct as our base model, chosen for its balance of capability and computational efficiency suitable for on-device deployment. Fine-tuning was performed using PEFT with LoRA adaptation, implemented through the Unsloth library for accelerated training and merging.

### 3.2 Dataset Preparation

Training data consisted of 1,500 synthetic medical transcriptions and corresponding structured medical notes specifically created for this project and available publicly [17]. This dataset focuses on endocrinology cases, providing domain-specific examples of physician-patient conversations and their corresponding clinical documentation.

The dataset was generated through an iterative, multi-stage data synthesis and refinement workflow. First, realistic endocrinology consultation topics were created, focusing on common disorders, symptoms, treatments, and lifestyle discussions typical in clinical visits. Detailed contextual descriptions were then developed for each topic to guide accurate and relevant transcript generation. Using advanced LLM prompting, natural and unedited endocrinology consultations were synthesized, incorporating labs, discussions, medication plans, and follow-up instructions while reflecting real conversational nuances such as interruptions and clarifications. Each generated transcript underwent an automated critique and revision loop, where completeness, clinical relevance, and realism were evaluated and iteratively improved based on feedback to enhance overall data fidelity. Finally, the refined transcripts were transformed into standardized, structured endocrinology notes formatted with key clinical sections typically seen in a structured medical note.

To validate the quality and clinical accuracy of our generation methodology, we initially produced 20 transcriptions and corresponding notes utilizing this workflow. These samples underwent evaluation by an endocrinologist to assess their medical accuracy, clinical realism, and adherence to appropriate documentation standards. Only upon receiving approval and confirmation that the generated content satisfied professional clinical standards did we proceed to expand the dataset to 1,500 samples, thereby ensuring the robustness and suitability of our generation methodology for large-scale application.

### 3.3 Evaluation Framework

Two distinct evaluation datasets were employed:

1. **Internal Evaluation Dataset:** 100 synthetic transcripts and structured medical notes specifically created for this project and publicly available [18].
2. **Modified ACI Benchmark:** 140 transcripts from ACI bench, with structured notes modified to match our training data format, also publicly available [19].

### 3.4 Evaluation Metrics

**Text Similarity Metrics:**

- ROUGE: We calculated ROUGE-1, ROUGE-2, ROUGE-L, and ROUGE-Lsum to measure n-gram overlap between generated and reference notes [20].
- BERTScore: This metric uses contextual embeddings from BERT to compute semantic similarity, providing precision, recall, and F1 scores [21].
- BLEURT: A learned metric that correlates well with human judgments, trained on synthetic data and human ratings [22].

**LLM-as-Judge Assessment:** Comprehensive evaluation across multiple dimensions was conducted using gpt-4.1-mini-2025-04-14. GPT 4.1 mini evaluated the LLM generated notes for following using a Likert scale like rating [23].

- Factual Correctness (1-5 scale)
- Completeness (1-5 scale)
- Clinical Relevance (1-5 scale)
- Coherence and Organization (1-5 scale)
- Negation Detection (binary)
- Terminology Accuracy (1-5 scale)
- Readability (1-5 scale)
- Overall Quality (1-5 scale)

**Clinical Safety Metrics:**

- Hallucination categorization (No, Minor, Major)
- Omission categorization (No, Minor, Major)

## 4. Results

### 4.1 Text Similarity Metrics Performance

Table 1 and figure 1 presents the text similarity evaluation results comparing the base Llama 3.2 1B model with our fine-tuned OnDevice model across both evaluation datasets.

**Table 1: Automated Evaluation Results**

| Dataset | Model | ROUGE-1 | ROUGE-2 | ROUGE-L | BERTScore F1 | BLEURT |
| --- | --- | --- | --- | --- | --- | --- |
| ACI Benchmark | Base_Llama | 0.346 | 0.118 | 0.199 | 0.832 | 0.447 |
| ACI Benchmark | OnDevice | 0.496 | 0.227 | 0.302 | 0.866 | 0.445 |
| Internal Eval | Base_Llama | 0.363 | 0.135 | 0.221 | 0.827 | 0.455 |
| Internal Eval | OnDevice | 0.653 | 0.390 | 0.471 | 0.907 | 0.516 |

The OnDevice model demonstrated substantial improvements across all metrics. On the ACI benchmark, ROUGE-1 scores increased by 43.3% (0.346 → 0.496), while ROUGE-2 improved by 92.7% (0.118 → 0.227). Even more dramatic improvements were observed on the internal evaluation dataset, with ROUGE-1 increasing by 79.9% (0.363 → 0.653) and ROUGE-2 by 188.5% (0.135 → 0.390). There was a slight decrease in BLEURT for OnDevice model 0.447 to 0.445. But for the internal evaluation dataset, there was improvement in BLEURT score from 0.455 to 0.516.

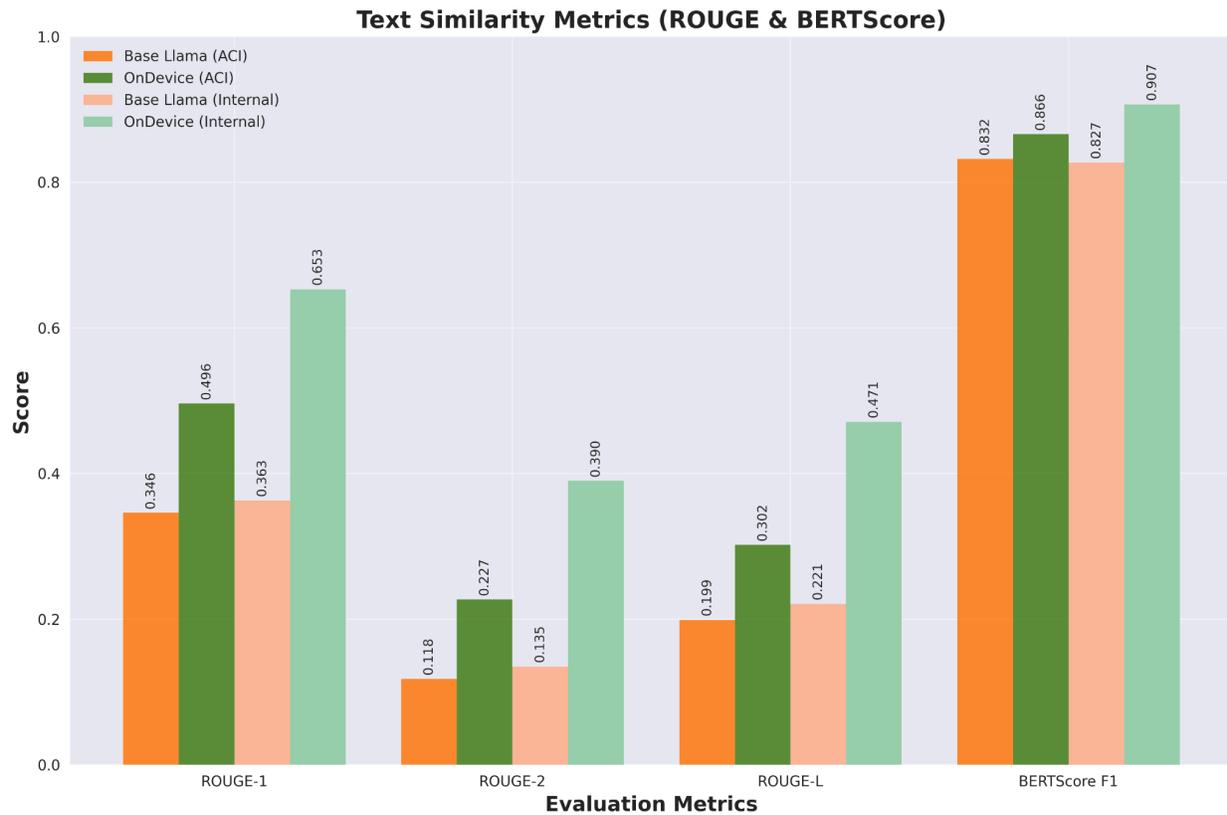

## 4.2 Clinical Quality Assessment

Figure 2 illustrates the LLM-as-judge evaluation results across multiple clinical quality dimensions.

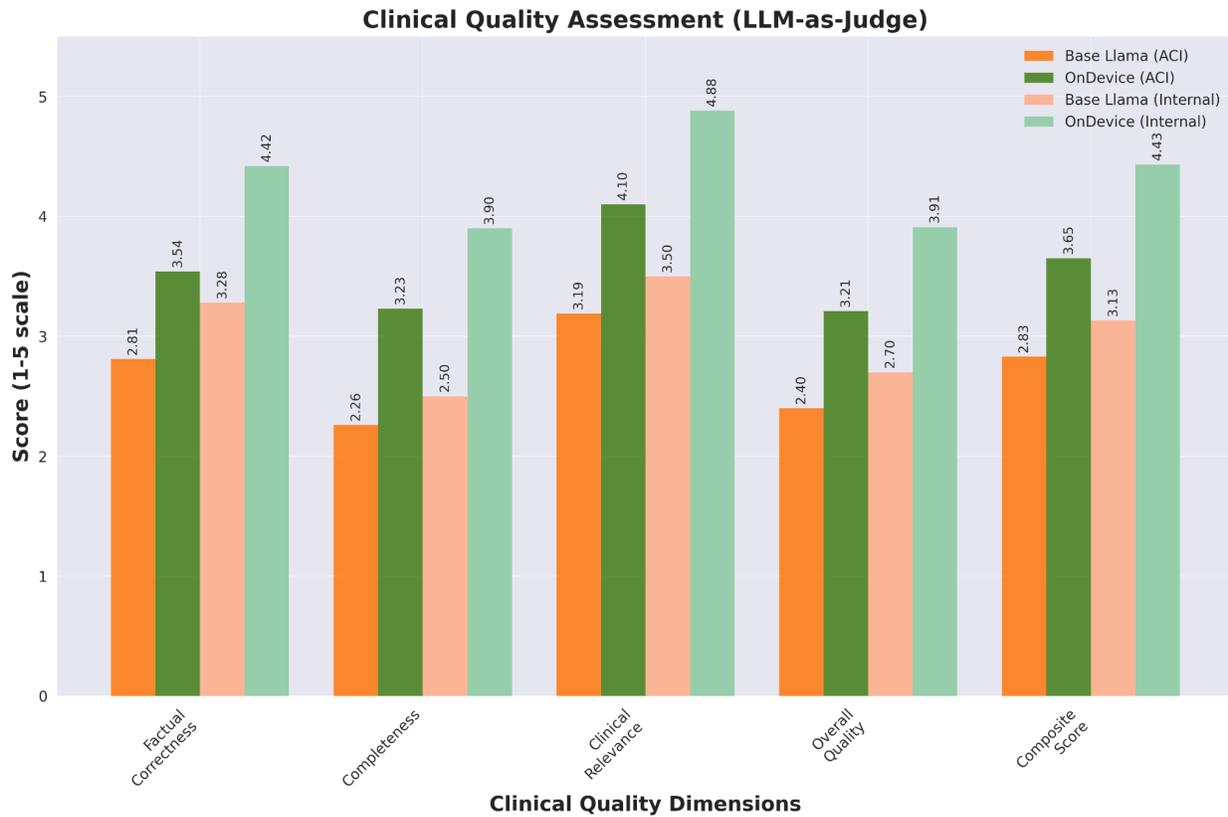

Table 2, presents the clinical quality evaluation results comparing the base Llama 3.2 1B model outputs with our fine-tuned OnDevice model across both evaluation datasets using GPT 4.1 mini as judge.

**Table 2: Clinical Quality Assessment Results**

| Dataset | Model | Factual Correctness | Completeness | Clinical Relevance | Overall Quality | Composite Score |
|---|---|---|---|---|---|---|
| ACI Benchmark | Base_Llama | 2.81 | 2.26 | 3.19 | 2.40 | 2.83 |
| ACI Benchmark | OnDevice | 3.54 | 3.23 | 4.10 | 3.21 | 3.65 |

| Dataset | Model | | | | |
|---|---|---|---|---|---|
| Internal Eval | Base_Llama | 3.28 | 2.50 | 3.50 | 2.70 | 3.13 |
| Internal Eval | OnDevice | 4.42 | 3.90 | 4.88 | 3.91 | 4.43 |

The fine-tuned model showed consistent improvements across all clinical quality dimensions. Most notably, factual correctness improved from 2.81 to 3.54 on the ACI benchmark and from 3.28 to 4.42 on the internal evaluation dataset.

## 4.3 Clinical Safety Analysis

Table 3 presents the hallucination and omission analysis, critical for clinical safety assessment.

**Table 3: Clinical Safety Metrics**

| Dataset | Model | No Hallucination | Minor Hallucination | Major Hallucination | No Omission | Minor Omission | Major Omission |
|---|---|---|---|---|---|---|---|
| ACI Benchmark | Base_Llama | 9 | 46 | 85 | 0 | 33 | 107 |
| ACI Benchmark | OnDevice | 23 | 82 | 35 | 2 | 117 | 21 |
| Internal Eval | Base_Llama | 11 | 56 | 33 | 0 | 29 | 71 |
| Internal Eval | OnDevice | 52 | 43 | 5 | 0 | 99 | 1 |

The OnDevice model demonstrated significant improvements in clinical safety. Major hallucinations decreased dramatically from 85 to 35 cases on the ACI benchmark (-58.8%) and from 33 to 5 cases on the internal evaluation dataset (-84.8%). Similarly, major omissions were substantially reduced from 107 to 21 cases (-80.4%) and from 71 to 1 case (-98.6%) respectively.

Figure 3 illustrates the number and type of hallucinations in outputs from each model for ACI and internal evaluation datasets.

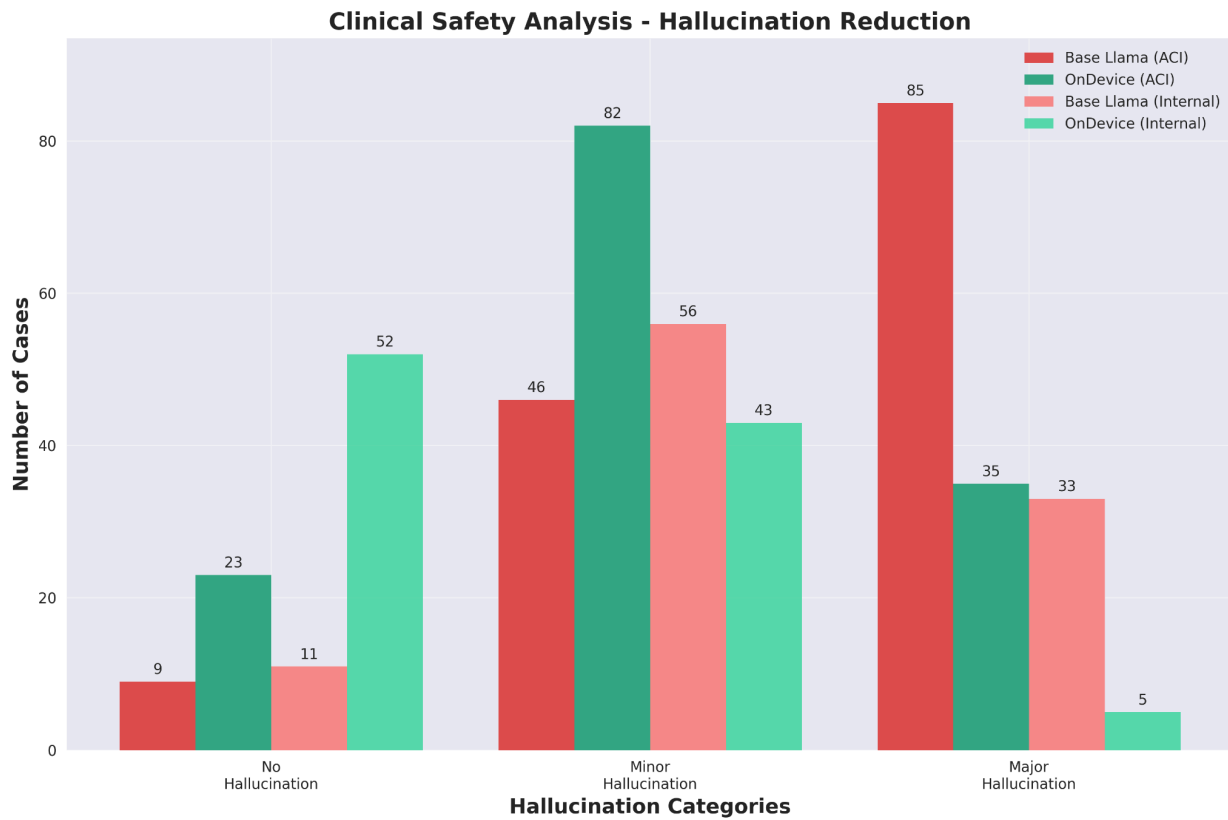

## 5. Discussion

Our evaluation demonstrates that the OnDevice model achieved consistent performance improvements across datasets, with particularly notable enhancements in clinical safety through reduced hallucinations and omissions. Given that studies estimate human-generated clinical notes have, on average, at least 1 error and 4 omissions, our focus on reducing hallucinations and omissions in automated transcription addresses a critical clinical need [24]. The model showed improved semantic understanding beyond lexical matching, and domain-specific training on endocrinology cases successfully translated to better performance on specialized evaluation tasks. The improvement in factual correctness and clinical relevance suggests that domain-specific fine-tuning can effectively adapt general-purpose models for specialized medical tasks.

The ability to achieve these improvements with a 1B parameter model is particularly significant for healthcare adoption. Traditional concerns about computational requirements and deployment costs are substantially mitigated, making the technology accessible to smaller practices and resource-constrained environments.

The on-device deployment model addresses fundamental privacy concerns in healthcare AI. By processing all data locally, the system eliminates risks associated with cloud-based processing while ensuring compliance with HIPAA and similar regulatory frameworks. This approach provides complete data sovereignty, allowing healthcare organizations to benefit from AI capabilities without compromising patient privacy.

Our work demonstrates several key technical contributions. The successful application of LoRA fine-tuning to medical transcription tasks shows that parameter-efficient methods can achieve substantial performance improvements with limited computational resources. Additionally, we developed a comprehensive evaluation framework that combines automated metrics with LLM-as-judge evaluation using GPT-4.1, providing a robust assessment methodology for clinical AI systems. Our emphasis on hallucination and omission analysis addresses critical concerns for medical AI deployment. To ensure reproducibility, we have made the evaluation code and prompts for the GPT-4.1 judge publicly available in the supplementary materials, enabling other researchers to replicate our assessment approach.

Several limitations should be acknowledged in this work. The evaluation focused primarily on endocrinology cases, and generalization to other medical specialties requires further validation. The training dataset of 1,500 examples, while sufficient for parameter-efficient fine-tuning, may limit the model's exposure to diverse clinical scenarios. Additionally, the LLM-as-judge evaluation, while comprehensive, may not capture all aspects of clinical utility that would be apparent to human clinicians. Finally, the evaluation was conducted on curated datasets and may not fully reflect the challenges of real-world clinical transcription environments.

Future research should address several key areas to advance this work. Expanding evaluation to multiple medical specialties will be essential to assess the model's generalization capabilities beyond endocrinology. Additionally, conducting real-world clinical trials with practicing clinicians will provide crucial insights into practical utility and workflow integration challenges. Finally, developing frameworks for continuous learning that enable ongoing model improvement based on clinical feedback while maintaining patient privacy will be critical for long-term deployment success.

## 6. Conclusion

This study demonstrates that fine-tuning compact large language models for structured medical note creation can achieve substantial improvements in clinical documentation quality while enabling complete on-device deployment. The OnDevice model showed consistent improvements across automated metrics, clinical quality assessments, and safety measures, with particularly notable reductions in major hallucinations and omissions.

The ability to achieve these results with a 1B parameter model addresses key barriers to AI adoption in healthcare: privacy concerns, computational requirements, and ongoing costs. By enabling local deployment performance, this approach makes advanced AI-powered clinical documentation accessible to healthcare organizations of all sizes.

The open-source release of the model, training data, evaluation framework, and browser-based deployment software provides a foundation for broader adoption and further research. This work contributes to the growing body of evidence that privacy-preserving, on-device AI can deliver clinically meaningful improvements in healthcare workflows while maintaining data security and patient privacy.

## Acknowledgments

The authors acknowledge the open-source community for providing the foundational tools and datasets that made this research possible. We thank the contributors to the Unsloth library and the HuggingFace ecosystem.

## Competing Interests

JT: No relevant disclosures
AM and WL: Owns [starfishdata.ai](starfishdata.ai), that generates synthetic data for model training.
NT: works at [alignmentlab.ai](alignmentlab.ai), an open-source AI research lab
ZM and DD: works at Solo Tech, a company that develops on-device AI models.

## Data Availability

- Model: https://huggingface.co/OnDeviceMedNotes/Medical_Summary_Notes
- Initial Verification Dataset https://huggingface.co/datasets/starfishdata/endocrinology_transcription_and_notes
- Training Dataset: [https://huggingface.co/datasets/starfishdata/endocrinology_structured_notes_1500](https://huggingface.co/datasets/starfishdata/endocrinology_structured_notes_1500)
- Evaluation Datasets:
    - https://huggingface.co/datasets/starfishdata/endocrinology_structured_notes_eval
    - https://huggingface.co/datasets/starfishdata/endocrinology_structured_notes_aci_bench_eval
- Open-source software for testing available at https://huggingface.co/spaces/Johnyquest7/medical-transcription-notes

## Funding

No external funding was received for this project.